\documentclass[runningheads]{llncs}
\usepackage[T1]{fontenc}

\usepackage{times}
\usepackage{latexsym}

\usepackage{graphicx}
\usepackage{url}
\usepackage{booktabs}
\usepackage{multirow}
\usepackage{amsmath}
\usepackage{bbm}
\usepackage{dsfont}
\usepackage{tikz}
 
\usepackage{pifont}% http://ctan.org/pkg/pifont
\newcommand{\cmark}{\ding{51}}%
\newcommand{\xmark}{\ding{55}}%

\usepackage{tcolorbox}
\usepackage{multicol}
\usepackage{lipsum}
\usepackage{verbatim}

\begin{document}
\title{Ontology-Constrained Generation of Domain-Specific Clinical Summaries}
%
% \titlerunning{Ontologies for Domain-aware and Constrained Generation from Clinical Notes}
% If the paper title is too long for the running head, you can set
% an abbreviated paper title here
%
\author{Gaya Mehenni \inst{1,2} \and
Amal Zouaq \inst{1,2}}
%
% \authorrunning{F. Author et al.}
% \author{Gaya Mehenni\inst{1}\orcidID{0000-1111-2222-3333} \and
% Amal Zouaq\inst{2,3}\orcidID{1111-2222-3333-4444}}
% %
% \authorrunning{F. Author et al.}
% First names are abbreviated in the running head.
% If there are more than two authors, 'et al.' is used.
%
% \and
% Springer Heidelberg, Tiergartenstr. 17, 69121 Heidelberg, Germany
% \url{http://www.springer.com/gp/computer-science/lncs} \and
% ABC Institute, Rupert-Karls-University Heidelberg, Heidelberg, Germany\\
% \email{\{abc,lncs\}@uni-heidelberg.de}
% \institute{LAMA-WeST Lab \and Polytechnique Montreal 
% }
\institute{LAMA-WeST \and Polytechnique Montreal\\
\email{\{gaya.mehenni,amal.zouaq\}@polymtl.ca}}
\maketitle              % typeset the header of the contribution
%

% \maketitle
\begin{abstract}
Large Language Models (LLMs) offer promising solutions for text summarization. However, some domains require specific information to be available in the summaries. Generating these domain-adapted summaries is still an open challenge. Similarly, hallucinations in generated content is a major drawback of current approaches, %especially in specialized domains, 
preventing their deployment. This study proposes a novel approach that leverages ontologies to create domain-adapted summaries both structured and unstructured. We employ an ontology-guided constrained decoding process to reduce hallucinations while improving relevance. When applied to the medical domain, our method shows potential in summarizing Electronic Health Records (EHRs) across different specialties, allowing doctors to focus on the most relevant information to their domain. %Additionally, structured summaries allow clinicians to easily query relevant information across clinical notes. 
Evaluation on the MIMIC-III dataset demonstrates improvements in generating domain-adapted summaries of clinical notes and hallucination reduction. %Overall, our work demonstrates the potential of LLM-based clinical summarization systems to reduce documentation burden and improve clinician efficiency.

\end{abstract}

\section{Introduction}
Large Language Models (LLMs) have shown major improvements in their extraction and summarization capabilities. In the medical field, these models offer the potential to automate the summarization process of complex medical data, such as Electronic Health Records (EHRs) and clinical notes \cite{searle_discharge_2023}. These documents, which contain an overwhelming amount of information, are a significant contributor to clinician burnout \cite{tajirian_influence_2020}. Thus, the generation of more focused and domain-specific summaries would help alleviate this task. Multiple challenges arise when LLMs are applied to the medical domain.
% Recent advancements in clinical summarization have shown the potential of automating tedious and time-consuming tasks, which could prevent clinician burnout \cite{sinsky_allocation_2016}. 
% Electronic Health Records (EHR) are a significant contributor to clinician burnout. Documentating every aspect of a patient's stay, these documents contain an overwhelming amount of structured and unstructured information. At the point of care, clinicians must extract relevant data through all this information to make an formal decision about the patient's diagnosis as well as writing a relevant discharge summary. This tedious and time-consuming task can lead to burnout \cite{sinsky_allocation_2016} and increase the risk of clinical errors \cite{salvagioni_physical_2017}. 
% While leveraging LLMs to automate this process can not only save time for clinicians, but also structure patient records. This structure can improve the readability of these documents and reduce this time-consuming task for doctors. However, multiple challenges arise when generating summaries of clinical notes using LLMs. 
Not only is the information condensed in a domain-specific terminology, but clinical notes are unstructured and do not follow specific formats. Additionally, these models are prone to hallucinations which can have serious consequences in healthcare settings. These issues become even more complex when the generated content must be adapted to different medical contexts. For example, the critical information required for cancer treatments differs significantly from that needed for diagnosis imaging analysis. Thus, ideally, different summaries should be generated for different areas of focus. To address these challenges, medical ontologies can be utilized to extract and prioritize information relevant to certain domains, specialties or fields. These ontologies provide a structured representation of medical knowledge, allowing for the identification of key concepts and relationships within a particular field aka \textit{domain}. This information can be used to extract relevant information from clinical notes and leverage it to produce domain-adapted summaries and %. Plus, since ontologies are based on the factual understanding of a domain, the instrinsic knowledge embedded in these structures can be utilized to 
to reduce hallucinations. In this context, a subsequent challenge is to constrain the generation of LLMs to specific domain concepts and properties and avoid the generation of non factual non-grounded information. 
Following the above, this paper focuses on three main research questions :
\begin{itemize}
    \item How can we adapt LLM-generated clinical summaries to different domains ?
    \item Can we leverage ontologies to constrain the generation of LLMs?
    \item How can we constrain the generation process of LLMs to reduce hallucinations ?
\end{itemize}
%To do so, we introduce new prompting and decoding algorithms based on ontologies. Using these algorithms, we constrain the LLM's output to align with the relationships and concepts defined in the ontology. Thus, we can reduce the likelihood of generating information that contradicts established domain knowledge, leading to more relevant answers and less hallucinated content. To the best of our knowledge, this is the first algorithm that employs ontologies in conjunction with LLMs to reduce hallucinations. Furthermore, combined to a pruning process, our method allows us to generate summaries that are more tailored to certain medical contexts while being more factual. Figure \ref{fig:acess} shows how our method can generate ontology-structured summaries of clinical notes and domain-adapted summaries. 
Our contributions can be summarized into the following aspects :
\begin{itemize}
    \item We show how ontologies can be harnessed by LLMs for a domain-aware and constrained generation; 
    \item We design a new ontology-guided decoding process that utilizes the knowledge embedded in an ontology to reduce hallucinations.  We constrain the LLM's output to align with the relationships and concepts defined in the ontology. Thus, we reduce the likelihood of generating information that contradicts established domain knowledge, leading to more relevant answers and less hallucinated content;
    \item We create a new summarization process that can generate domain-tailored summaries of clinical notes.
\end{itemize}

\begin{figure*}[h] 
    \centering
    \includegraphics[width=\linewidth]{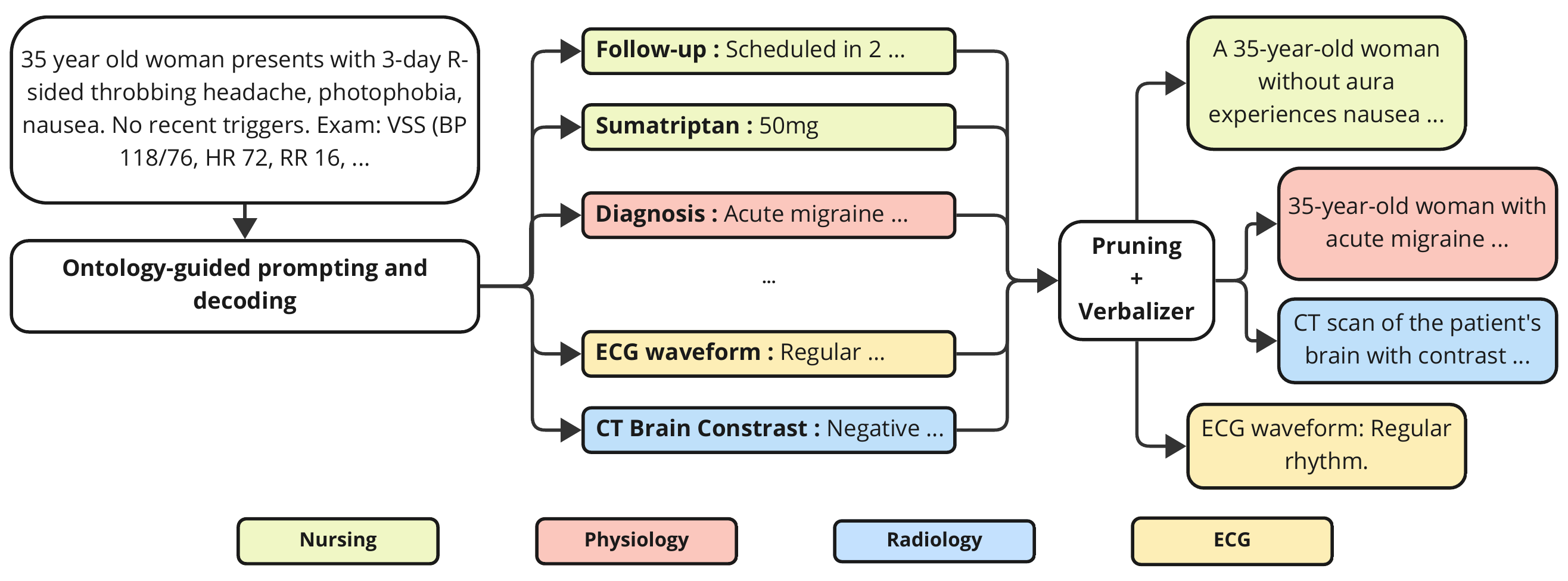}
    \caption{Overview of our general architecture to generate domain-tailored summaries. }
    \label{fig:domain_acess}
\end{figure*}

To the best of our knowledge, our work is the first approach that employs ontologies in conjunction with LLMs to constrain generation. Figure \ref{fig:domain_acess} shows how our approach can generate ontology-structured summaries of clinical notes and domain-adapted summaries \protect\footnote{Code is available at https://github.com/Lama-West/Ontology-based-decoding\_EKAW2024.}.

\section{Related work}

\paragraph{Summarization.}
Summarization is the process of generating a smaller text from a larger input text. The main objective of that process is to grasp the information of the input text in the most condensed way possible. Existing work on the task of summarization can be divided into two main groups : \textit{extractive} and \textit{abstractive} summarization. Extractive summarization involves retrieving sections where the information is mostly present \cite{logan_iv_baracks_2019,liu_deakinnlp_2023}. Abstractive summarization relies on neural architectures to generate a reformulated version of the original text. While the latter usually works better due to its context-dependent summarization properties, recent work have tried utilizing a hybrid approach \cite{van_zandvoort_enhancing_2024,van_veen_adapted_2024}. These works usually obtain better performance \cite{shi_di_2022} since they guide abstractive models towards relevant sections of the input while keeping the advantages of extractive methods. Our work, which is also based on this hybrid approach, adds an ontological component to the abstractive and extractive modules to improve the final summarization capabilities.

\paragraph{Clinical Text Summarization. }
The development of clinical note summarization is especially slow due to the absence of reference summaries \cite{searle_discharge_2023}. Prior work on clinical summarization has mainly focused on summarizing radiology reports \cite{van_veen_clinical_2023,chuang_spec_2023} and generating sections of the discharge summary of an admission \cite{searle_discharge_2023,adams_speer_2024}. In contrast to current methods that simply generate an output text, we aim to generate a more structured report tailored to the needs of a given specialist. This is particularly useful as clinical narratives often exhibit significant variation in how the information is structured and presented \cite{sorita_ideal_2021}. This would allow the clinician to quickly query on certain attributes of the patient while having access to the full report if needed. In this study, we rely on a medical ontology to structure the summaries.

\paragraph{Hallucination. }
One problem that arises from current summarization techniques is that LLMs are still prone to hallucinations \cite{guerreiro_hallucinations_2023}, slowing down their deployment in the medical domain. Multiple techniques have been developed to mitigate this problem. They can be regrouped into design-time solutions, training-time solutions, generation-time solutions and external tools. Design time solutions include modifying the base architecture of LLMs to reduce hallucinations \cite{gao_leveraging_2023}. Training-time solutions include reinforcement learning \cite{wu_fine-grained_2023}, latent space understanding \cite{li_inference-time_2023} and loss function modification \cite{chern_improving_2023}. As for generation-time solutions, they include pre-generation prompting techniques \cite{wei_chain--thought_2022,press_measuring_2023} and post-generation evaluation \cite{dhuliawala_chain--verification_2023,cohen_lm_2023}. The latter measures the confidence of a model on its generation. Finally, external tools solutions incorporate search engines and vector databases into the loop \cite{lewis_retrieval-augmented_2020}. In this work, our contribution is to propose  ontologies to identify relevant beams during generation to improve factuality.

\paragraph{Constrained Generation. }
Constrained generation has seen significant advancements in recent years, addressing the need for generated content to follow specific constraints. These constraints can, for example, be utilized to impose a certain structure in the generation process. While some \cite{willard_efficient_2023} have utilized finite state machines to generate a structured output, others \cite{zheng_sglang_2024} improved the efficiency of the process by fully exploiting the constraints to reduce the number of inference passes. \cite{stengel-eskin_zero_2024} have also applied constraints for a structured output through grammars, making their method even more modular. \cite{geng_grammar-constrained_2023} extended this process with a framework allowing grammars to be input-dependent. %, %to the context, making the output structure dependent on the input. 
Lexically constrained generation methods often rely on logits re-weighting or beam pruning to guide the model towards certain words or concepts \cite{yang_fudge_2021}. Recent work have improved these methods through the use of heuristic estimates \cite{lu_neurologic_dec_2021} and knowledge graphs \cite{choi_kcts_2023}. To our knowledge, very few, if any, state-of-the-art approaches have leveraged ontologies for constrained generation. Our work is a step in this direction. 

\section{Methodology}

\begin{figure*}[h] 
    \centering
    \includegraphics[width=\linewidth]{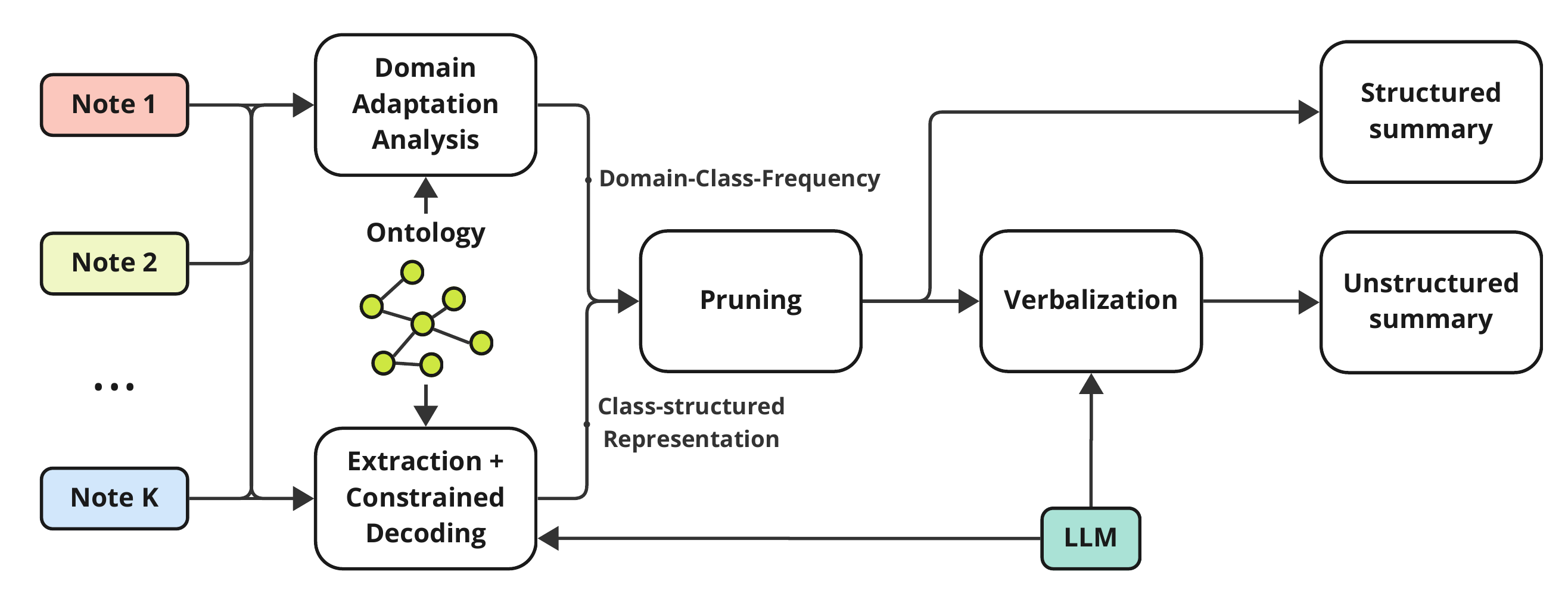}
    \caption{\textbf{Overall architecture of our method : } Multiple notes about the same patient are passed to the framework and structured and unstructured summaries are generated}
    \label{fig:acess}
\end{figure*}

Our research explores the potential of using ontologies to guide a language model towards relevant information using prompting and constrained generation. By constraining the generation using ontological structures, we aim to improve the summarization capabilities of language models and to reduce their hallucinations. To do this, we propose to utilize the ontology in conjunction with the beam search algorithm to assess the relevance and factual accuracy of potential beam candidates in relation to the input. By implementing this ontology-guided beam search, we expect to enhance the overall coherence and reliability of the generated text, ensuring that it aligns more closely with the knowledge represented in the ontology. To reduce hallucinations, we propose to also evaluate the beam paths based on the clinical note to favor those that resemble it the most. 
As a proof of concept, we develop a new method which divides the summarization task into multiple simple inference passes guided by an ontology-based prompting approach. Figure \ref{fig:acess} shows how, given multiple clinical notes about the same patient, our method outputs a text summary and a structured summary whose structure is defined by ontology concepts. %The structured approach enhances the readability and utility of the summaries, allowing healthcare professionals to quickly grasp and query essential information. 
%By organizing the extracted data into a structured format, our approach ensures that critical details are highlighted and preserved, thus improving the overall effectiveness of the summarization process. 
This structured representation of the clinical notes can be leveraged afterwards to adapt the final summary to various domains (medical fields such as cardiology, oncology, etc.). Finally, our approach can be applied to any model since it only requires token probabilities.

\subsection{Domain adaptation analysis}\label{sec:domain_analysis}

We define a domain to be a set of ontology classes of interest related to a specific medical field. To adapt the generation to multiple domains, an initial analysis is performed using texts from  different domains to identify their key concepts. Given texts that are linked to a certain domain $D$, we identify important concepts, or classes, in each text using an ontology-based annotator and create a set $S$ based on a minimum occurrence threshold. We presume that the annotator can detect different formulations of the same concept (abbreviations, plurals, etc). Then, using the ontology, we retrieve all ancestors of each class and add the ancestors to $S$. Subsequently, the frequency of each class in $S$ is computed and stored in a class-to-frequency dictionary. We define this dictionary as the Domain-Class-Frequency dictionary (DCF). Its goal is to store the most relevant classes in a domain to guide the generations towards these classes. Then, each DCF is normalized according to the average DCF, computed by averaging the classes frequencies of all domains. This normalizing step allows the DCF to only contain relevant classes to the domain and reduce the weight of general medical concepts (which are more frequent). The DCF stores the important domain concepts to specialize the summary at the next stage.

\subsection{Information Extraction using Ontology-based Prompting}\label{sec:extraction}
\begin{figure}[h] 
    \centering
    \includegraphics[width=\linewidth]{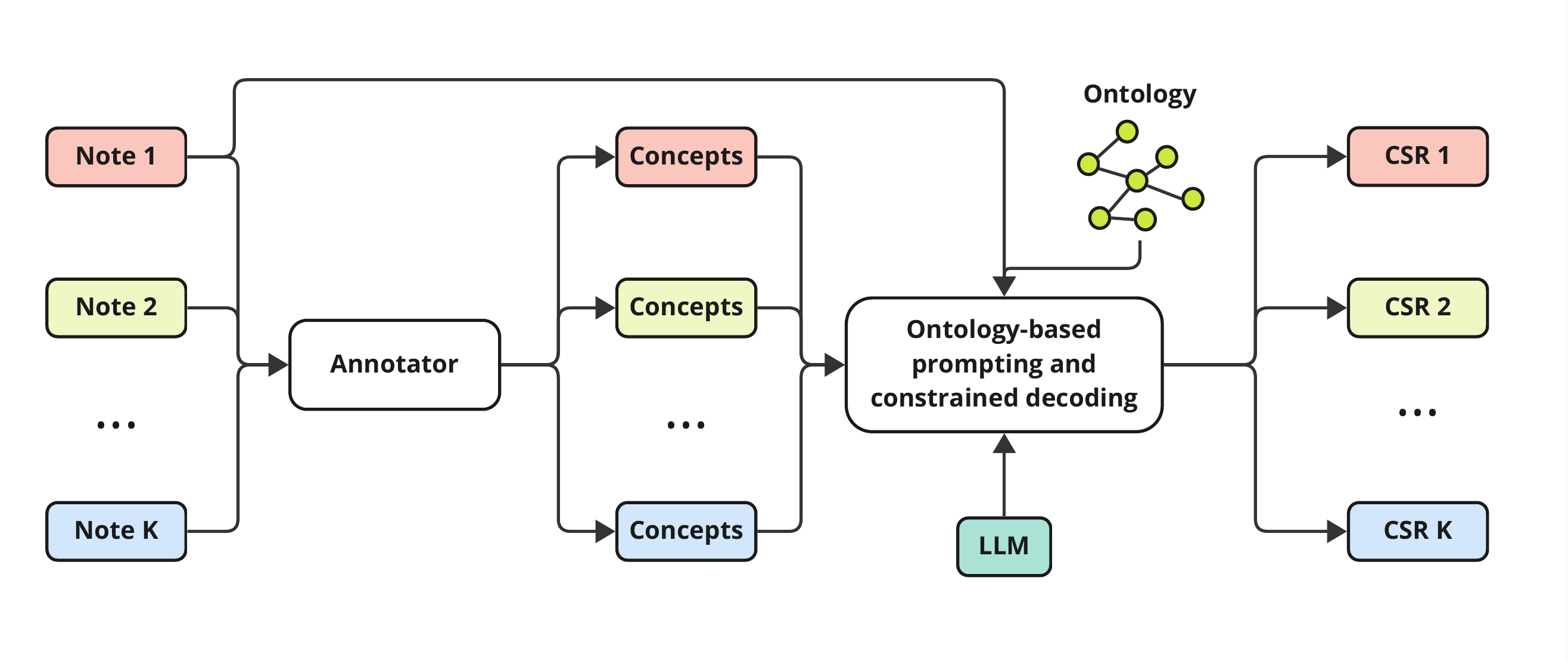}
    \caption{Extraction Phase}
    \label{fig:faa}
\end{figure}

Our next step is to extract medical concepts, properties and their values from clinical notes about a patient using a large language model (LLM), an ontology-based prompting process and a constrained decoding strategy. The overall process of the extraction phase is shown in Figure \ref{fig:faa}. The main goal of this step is to generate a structured version of clinical notes allowing doctors to easily query information based on concepts. This structured version will also be used in conjunction with the DCF dictionary (see section \ref{sec:domain_analysis}) in latter steps to specialize the summary to a given domain.

To improve the information extraction capabilities of the LLM, we adapted \cite{wang_element-aware_2023}'s summarization technique to the medical domain by incorporating an ontology-guided prompting process. Given multiple clinical notes of a patient during an admission and a medical ontology, we start by annotating each clinical note %one by one 
to retrieve all the medical concepts mentioned in each note. This process can easily be parallelized since, at this step, each note is considered independently. We then prompt the model to summarize each note around the concepts in multiple inference passes. Following a RAG-like \cite{lewis_retrieval-augmented_2020} approach, we also augment the prompt with all relevant restriction properties that can be inferred using the ontology class. This augmentation provides the model with more context about what the ontology class is referring to. For instance, if the model does not know what an electrocardiogram is, augmenting the prompt with characteristics of this concept can improve its extraction capabilities since a certain definition is given. The following prompt is used in that case :\\

\textit{Here is a clinical note about a patient : [clinical note]. In a short sentence, summarize everything related to the "[concept]" concept mentioned in the clinical note. "[concept]" is characterized by [properties]. If nothing is mentioned, answer with "N/A"}. \label{prompt_template}

For example, the prompt applied to the "electrocardiogram" concept would be : \\

 \textit{Here is a clinical note about a patient : [clinical note]. In a short sentence, summarize everything related to the "electrocardiogram" concept mentioned in the clinical note. "electrocardiogram" is characterized by Evaluation - action AND Heart Structure AND Electrocardiographic monitor and recorder, device. 
 If nothing is mentioned, answer with "N/A"}. \\

%An example of a prompt is given in the appendix, in section \ref{appendix:prompt_examples}. 
Then, for each clinical note, a class-structured representation (CSR) is created. The CSR is a mapping of all ontology classes (concepts) detected in the note to short summaries - extracted from the notes - of the information related to each class. These summaries, which we refer to as "extracted values", are generated by a model using the mentioned prompt template.\label{extracted_values}

% we start by annotating the clinical notes with the main medical concepts. Then, we retrieve the $n$ most frequent concepts and extract  their relevant properties using the associated ontology classes. We then prompt the model to summarize the notes around those base concepts using the extracted properties with the following prompt format :

\subsection{Constrained Decoding}\label{sec:constrained_decoding}
\begin{figure*}[h] 
    \centering
    \includegraphics[width=\linewidth]{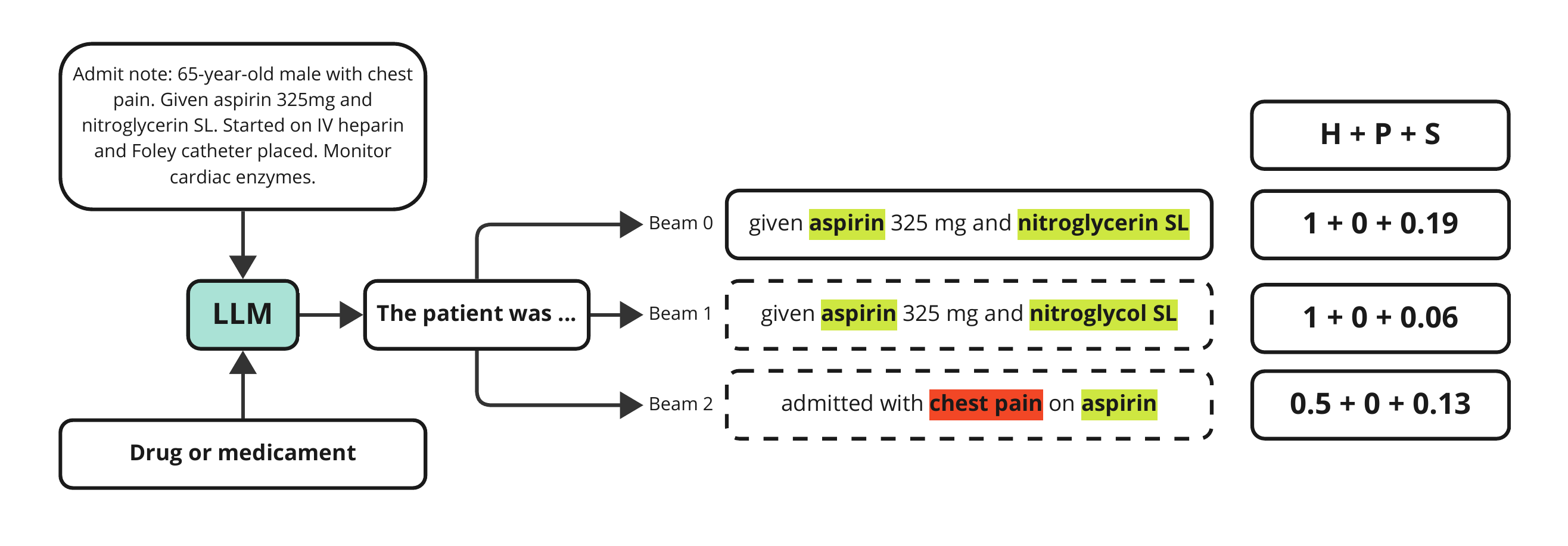}
    \caption{\textbf{Constrained decoding process : } Each beam rectangle represents the current generation window associated with the beam. The concepts in green are concepts that are associated to a children class of the base class (Drug or medicament) in the ontology. Green concepts improve the hierarchy score which augments the probability that the beam will be chosen as a final output. The similarity score is computed using the ROUGE-2 score between the generation window and the clinical notes.}
    \label{fig:constrained_decoding}
\end{figure*}

The decoding strategy during the extraction step is designed to guide the model's answer towards responses that are more relevant to the prompt. This strategy serves two primary purposes: leveraging the knowledge embedded within the ontology to steer the model towards a more relevant answer to the prompt (\textit{relevance)}, and reducing the occurrence of hallucinations (\textit{groundedness}) during the generation process to ensure that the model's responses align with both the prompt and the provided notes.  The overall constrained decoding process is shown in Fig. \ref{fig:constrained_decoding}. %As shown in Figure \ref{fig:acess}, 
This process is used in conjunction with the ontology-based prompting (see section \ref{sec:extraction}) to obtain a final class-structured representation (CSR) for each clinical note.

We propose an algorithm based on diverse beam search \cite{vijayakumar_diverse_2018}, wherein grouped beam search is employed to diversify the results. In a nutshell, our algorithm favors beams that textually resemble the input and that contain concepts that are related in the ontology through hierarchical relations(subclasses and superclasses) and restriction properties. Since extracting information from generated beam candidates is not an operation that can be done trivially token by token, our algorithm computes the beam scores after a certain number of tokens, defined as the \textit{generation window}. After this threshold, the newly generated tokens are analyzed using the same annotator used to tag the clinical notes. %Without the generation window, this tagging would not be possible easily as the annotator would not be able to tag anything in a single token sequence. 
%By defining this generation window, we attenuate the overhead created by usual beam search based algorithms \cite{hokamp_lexically_2017,saxena2023prompt} where logits are modified token-by-token to favor certain paths. Finally,
Once the beam is tagged, a score is calculated, as detailed below, %to reduce hallucinations and 
to favor beam paths that are aligned with the internal structure of the ontology and with the note content. 
% Every generation window, defined as the number of tokens the model is allowed to generate before our algorithm modifies the beam scores, the newly generated tokens are analyzed using the same annotator used to tag the clinical notes. A score is then calculated, as detailed below, to ensure a maximum overlap between... The generation window is set to a specific size because extracting information from the generation to compare it to the clinical notes and the ontology is an operation that cannot be done trivially token by token. Furthermore, 

To calculate the score of a beam, we compute three sub-scores : the hierarchy score $H$, the property score $P$ and the similarity score $S$. For all scores, we define $b$ to be the base class corresponding to the ontology class used in the prompt shown in section \ref{prompt_template}. This corresponds to the ontology class replacing the "[concept]" tag. We also define $T$ to be the newly generated tokens in the beam, $C$ the set of classes retrieved from $T$ and $A(c)$ the set of ancestors of a class $c$ in  $C$.

\paragraph{Hierarchy Score. } The hierarchy score $H$ computes a score based on the number of descendants of the base class that are present in the generated beam : 
\begin{equation*}
    H = H_{bf} \frac{1}{|C|} \sum_{c \in C} {\mathds{1}}\{ b \in A(c) \} 
\end{equation*}
where $H_{bf}$ is the hierarchy boost factor, a hyperparameter controlling how relevant we want the hierarchy score to have an impact on the final beam score. The primary goal of the hierarchy score is to guide beams towards the ontology's hierarchy to make sure that the generation is relevant to what is expected. For instance, when asked to summarize the patient's diseases, we would expect the model's answer to contain ontology classes that inherit the disease class.

\paragraph{Property Score.} The property score $P$ evaluates how relevant a beam is to the base class, based on restriction properties associated to the base class. While this can be generalized to any class property, we only consider restriction properties as they are the most frequent in the ontology used in our case. This score allows the decoding process to incorporate knowledge about concepts that can be inferred from the ontology. We thus want to favor beams that mention classes present in the restriction property values. For example, when asked about the "Fever" concept, we want the model's answer to ideally specify, if mentioned in the notes, that the patient has a body temperature above the normal range. We would then favor beams that mention the "Body Temperature" class. Following this intuition, the property score is given by :
\begin{equation*}
    P = \frac{P_{bf} \sum_{c \in C}{\mathds{1}}\{ c \in P(b) \}}{|C| |P(b)|}  + \text{R2}(T, P^{'}(b))
\end{equation*}
where $P_{bf}$, similar to $H_{bf}$, is the property boost factor and R2 is the ROUGE-2 score. $P(c)$ is the set of classes related to $c$ through restriction properties and $P^{'}(c)$ is a natural language formulation of $P(c)$. In practice, we only take into account \textit{And} and \textit{Or} restrictions. Given an object property restriction of the form \{ \textit{property1} : \textit{value1}, ... \}, to compute $P^{'}(c)$, we simply concatenate all values in the case of an \textit{And} restriction. For example, if $P(\text{Fever})$ = \textit{AND(Interprets: Body Temperature, Has Interpretation: Above Reference Range)}, then $P^{'}(\text{Fever})$ = \textit{Body Temperature Above Reference Range.} In the case of an \textit{Or} restriction, we add \textit{or} between every value. Computing the ROUGE-2 score between the natural language formulation of the restrictions and the newly generated tokens allows us to favor the beam if the annotator did not tag concepts correctly. 
\paragraph{Similarity Score.} 
Finally, the similarity score $S$ aims to measure how similar a beam is to the clinical notes. Since the model's answer about a single class should be short and resemble the clinical notes in an extractive manner, we hypothesize that the ROUGE-2 score is a good measure of the similarity of the answer to the notes. Its formula is simply given by :
\begin{equation*}
    S = S_{bf}\; \text{R2}(b, N)
\end{equation*}
where $S_{bf}$ is the similarity boost factor, $N$ is the clinical note and $b$ the current beam. 
\paragraph{Overall Score.} The final beam score $BS$ is given by :
\begin{equation*}
    BS = \text{LogSoftmax}(H + P + S)
\end{equation*}
\subsection{Pruning and Verbalization}
\paragraph{Pruning.} Once a class-structured representation of each clinical note is obtained using the extraction process, we prune each representation to adapt the summary to a given domain (or specialty) by keeping only the domain-relevant ontological classes. This pruning step aims to adapt the summaries to a given domain by focusing only on the extracted information that is relevant to the domain. This is done using the DCF dictionary computed initially (see section \ref{sec:domain_analysis}). In practice, for computational reasons, we  retrieve the top-$k$ most frequent classes in the DCF dictionary. However, %due to the way the DCF dictionary is computed, 
upper-level classes tend to be more frequent. To account for this problem, we also retrieve all classes that are within $\alpha$ nodes from a frequent class in the ontology using hierarchical relationships from superclasses to subclasses. In this case, $k$ is a hyperparameter controlling the length of generated summaries in terms of the number of concepts covered and $\alpha$ is a hyperparameter controlling the preciseness of the generated summary.

% \subsection{Merging}
% After the pruning step, we obtain, for each clinical note, a domain-adapted class-structured representation. The merging process's goal is to transform the sequence the representations associated to all notes to a single larger representation. Since the same information can be present multiple times in clinical notes, this step also removes redundant entries. It is designed to create a comprehensive summary that integrates all notes into a single clinical note, potentially improving the overall quality and coherence of the final summary. We achieve this by adding different concepts from multiple notes to the larger representation. For concepts that are present in multiple notes, their values are appended if they are similar. Otherwise, we keep the longer value. Their similarity is determined by the number of overlapping concepts in their respective values.

\paragraph{Verbalization.}
%Our process is designed to produce a structured representation of clinical notes. However, certain tasks require this information to be presented as an unstructured textual summary. To address this, we employ a 
We also use a final forward pass using the same LLM that transforms the structured output into an unstructured format. This process ensures that the output aligns with the requirements of the task and allows us to compare the efficiency of the method in terms of summarization. 

\section{Experiments}

\subsection{Dataset \& Ontology}

The Medical Information Mart for Intensive Care (MIMIC-III) database \cite{johnson_mimic-iii_2016} was used to retrieve clinical notes. This database regroups over 45,000 de-identified patient admissions to critical care in the Beth Israel Deaconess Medical Center between 2001 and 2012. Each admission contains multiple clinical notes associated to the same patient and is linked to a Brief Hospital Course (BHC) section of the patient's discharge summary. Overall, the dataset contains over 1.4 million single clinical notes. For computational reasons, we only use a test set called $\Omega$ of 400 randomly chosen admissions, regrouping over 3000 clinical notes, to perform our evaluations. \\

\begin{figure*}[h]
% \minipage{0.5\textwidth}
    \centering
    \includegraphics[width=0.5\linewidth]{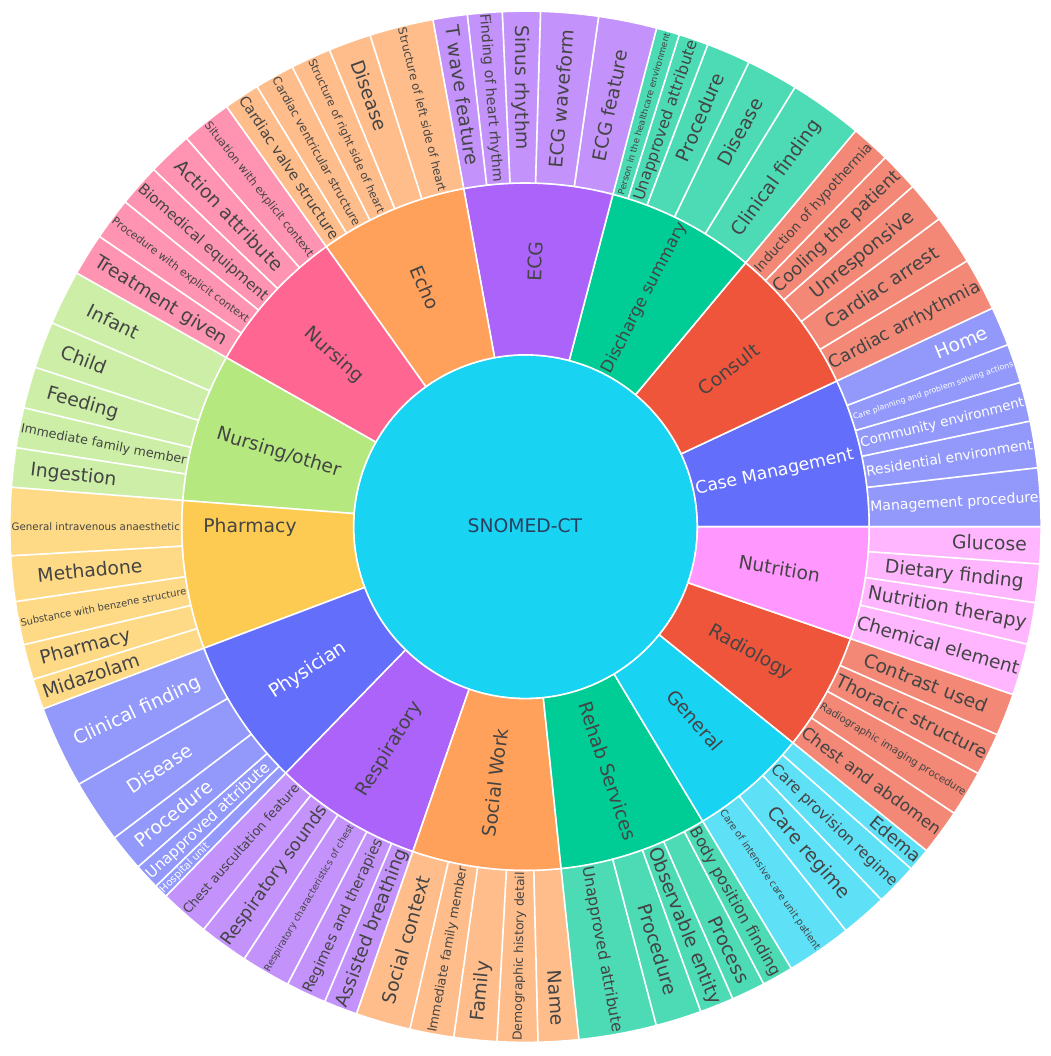}
% \endminipage\hfill
% \minipage{0.5\textwidth}
  % \includegraphics[width=\linewidth]{ACESS/concept_flow_echo.png}
% \endminipage\hfill
\caption{\textbf{Most frequent classes in SNOMED-CT based on different domains of clinical notes}. This plot was computed by automatically annotating 1000 clinical notes from each category in MIMIC-III and associating each concept to a SNOMED-CT class. We then performed a domain adaptation analysis as shown in section \ref{sec:domain_analysis}. Only the 5 most frequent concepts were kept in this figure.}
\label{domain_adaptation_analysis_mimic}
\end{figure*}

Clinical notes are associated to several categories ranging from electrocardiogram (ECG) reports to nursing notes and discharge summaries. Each note category serves a specific purpose and is associated to its own set of medical terms. For instance, ECG reports offer detailed insights about the patient's cardiovascular functions and structures, and nursing notes provide a continuous narrative of the patient's day-to-day care. These categories (ECG, Nursing, ...) are used to define our domains. For the domain adaptation of summaries, we employed the SNOMED-CT ontology \cite{Stearns2001-zn-snomed}, given its comprehensive representation of diverse medical fields. To analyze the occurrence of the SNOMED concepts in the MIMIC dataset, we computed the most frequent concepts in each domain (category) of MIMIC's notes.  As shown in Fig.\ref{domain_adaptation_analysis_mimic}, the domains differ by a lot on their class representation. 
Note that to perform this analysis, we pruned some branches of the SNOMED-CT ontology as they do not correspond to medical concepts (Linkage concepts, Qualifier values, ...).  %This subset, which we call $\Omega$, is used in all our experiments.

\subsection{Models}
To generate our tailored summaries, we used the Phi-3 model, a 3.5 billion parameter model matching GPT3.5's performance \cite{abdin_phi-3_2024} as well as Zephyr-7b-beta \cite{tunstall_zephyr_2023}. Both models were used for the extraction phase and as verbalizers. %using the same quantization config (see \ref{appendix:quantization}). 
As for the annotator linking text sequences and SNOMED-CT classes, we utilized the MedCAT annotator \cite{medcat_Kraljevic2021-ln}. This annotator was used during the decoding process and during evaluation to tag concepts (see Section \ref{sec:eval_summarization}). %an LSTM-based named-entity recognition model.
\paragraph{Generation window.} We evaluate empirically that a generation window of 5 to 15 tokens is a good trade-off between extraction capabilities and performance. We use a generation window of 10 tokens.
\paragraph{Hyperparameters.} For the constrained decoding process, we use $H_{bf} = 3, P_{bf} = 10, S_{bf} = 10$, a beam size of 10 and a group size of 2 for diverse beam search. As for the pruning phase, we use $k = 30$ and $\alpha = 2$. 

\section{Evaluation}

\subsection{Domain adaptation}
 \paragraph{Evaluator.} Since no ground truth exists for this task, we needed to train an evaluator to compute a domain adaptation score. For this purpose, we fine-tuned a BERT classifier trained on the medical domain \cite{alsentzer_publicly_2019} to predict the domain of a clinical note. We denote this model as the evaluator model. To do so, we utilized the \textit{CATEGORY} column of MIMIC-III which indicates the medical field of a clinical note. In practice, since not all domains are equally present, we focus on the \textit{Nursing}, \textit{ECG}, \textit{Radiology} and \textit{Physician} domains as they are the most frequent. To train the model, we use 400k individual clinical notes with their respective categories. For validation, 40k clinical notes were randomly selected. These training and validation sets are disjoints from the test set used to evaluate the summaries in other experiments. After training, the model achieved a 99\% accuracy on the validation set. However, we found that the model tended to rely on each domain's note format instead of relying entirely on the underlying concepts to predict the domain. Thus, we further fine-tuned the model on a custom-made dataset generated by passing 4k clinical notes not seen during the initial training through a BART paraphraser.
\paragraph{Domain Score.} We compute the domain score $D$ as the average logit score of the expected domain :
\begin{equation}
    D = \frac{1}{N}\sum_{i=0}^{N} \texttt{EVALUATOR}(d(x_i))[d_i]
\end{equation}
where $d(x_i)$ is the domain adapted summary of the admission $x_i$, $d_i$ is the expected domain and $N$ is the number of samples. This metric is used to evaluate how tailored the generated summary is based on the domain. 
\paragraph{Baselines.} To assess the effectiveness of domain adaptation, we evaluated three distinct approaches: (1) greedy search, (2) diverse beam search and (3) our method. For the greedy and diverse beam search methods, we augment the prompt with a prefix specifying the desired domain, instructing the model to tailor the summary accordingly\protect\footnote{See the GitHub repository for a prompt example}. 
%The prompt format is given in section \ref{appendix:domain_adaptation}. 
For our method, we simply pass the result of the pruning phase to the verbalizer without specifying the desired domain. We then pass each of the summaries (greedy, beam search, our method) through our BERT Evaluator. By comparing these three methods, we aim to demonstrate the relative efficiency of our proposed technique in producing domain-tailored summaries without the need for explicit prompt engineering. We evaluate these methods on the $\Omega$ test set. Since each admission can be used to generate 4 domain-adapted summaries (\textit{Nursing}, \textit{ECG}, \textit{Radiology} and \textit{Physician}), the final test set, in this experiment, contains 1600 pairs (admission, domain-adapted summary). 
\begin{table*}[h]
    \centering
    \begin{tabular}{lcc}
    \toprule
    Method              & \;\;\;\;Phi-3\;\;\;\;         & \;\;\;\;Zephyr\;\;\;\;        \\ 
    \midrule
    Greedy Search              & 0.41          & 0.56          \\ 
    \hline
    Diverse Beam Search & 0.43          & 0.59          \\ 
    \hline
    Ours (Extraction+Pruning)               & \textbf{0.86} & \textbf{0.78} \\ 
    \bottomrule
    \end{tabular}
    \caption{Domain score of generated summaries using greedy search, diverse beam search and our method}
    \label{tab:domain_score}
\end{table*}

Table \ref{tab:domain_score} shows the domain score of greedy and diverse beam search summaries versus our approach. The results highlight an improvement in domain adaptation when using our method. %, achieving an increase of over 39 \% with both models. 
In the case of Phi-3, it achieves more than twice the domain score of greedy search. 

\subsection{Hallucination Reduction}
\paragraph{Evaluator. } We also focus on evaluating the impact of our proposed constrained decoding process on reducing hallucinations. More precisely, we aim to evaluate how the constrained decoding process improves the extraction capabilities of the model. We evaluate two aspects : the \textit{groundedness} and the \textit{relevance} of the answers. We formulate this problem as an entailment task where we leverage Natural Language Inference (NLI) models to determine the entailment between our results and the clinical notes.  We ran this experiment on 1000 randomly extracted values of the $\Omega$ test set and used a DEBERTA model \cite{he2021deberta} fine-tuned on the MNLI dataset \cite{mnli-1101} as the NLI model.
\paragraph{Groundedness. } For an admission, we apply the extraction phase (see section \ref{sec:extraction}) using all methods, resulting in multiple CSRs (one per clinical note). In a CSR, each extracted value prepended by its concept ([concept] : [extracted value]) is considered as a hypothesis of the clinical note (the premise). We then compute the groundedness score by taking the entailment score as determined by an NLI model. The final groundedness score is calculated by averaging over all CSRs. 
\paragraph{Relevance score. } We also compute a relevance score where keys from a CSR (concepts) are considered the hypotheses and the extracted values the premises. Since the keys are used to generate the prompt question, computing this score allows us to evaluate how relevant the answer was based on the concepts.
\paragraph{Baselines.} Similar to the domain score, we employ a comparative analysis of three distinct generation methods during the extraction phase to quantify the effectiveness of our approach: (1) greedy search, (2) diverse beam search, and (3) our novel constrained generation. 
\begin{table*}[h]
    \centering
\begin{tabular}{lcccccc}
\toprule
                    & \multicolumn{3}{c}{Phi-3}                     & \multicolumn{3}{c}{Zephyr}         \\
Method              & \;\;\;\;Gnd\;\;\;\;  & \;\;\;\;Rel\;\;\;\;     & 
\;\;\;\;Avg\;\;\;\;       & \;\;\;\;Gnd\;\;\;\; & \;\;\;\;Rel\;\;\;\; & \;\;\;\;Avg\;\;\;\; \\
\midrule
Greedy Search   & 0.83          & 0.57          &
0.70          &              0.83&           0.82&    0.83     \\ \hline
Diverse Beam Search & 0.85          & 0.58          & 0.72          &      0.87        &     0.83      &    0.85     \\
\hline
Constrained Generation               & \textbf{0.90} & \textbf{0.64} & \textbf{0.78} &        \textbf{0.90 }     &     \textbf{0.88}      &    \textbf{0.89}    \\
\bottomrule
\end{tabular}
    \caption{Groundedness (Gnd) and Relevance (Rel) of extracted sequences using greedy search, diverse beam search and our ontology-constrained generation}
    \label{tab:groundedness_score}
\end{table*}

Table \ref{tab:groundedness_score} presents the groundedness and relevance scores for different generation methods. The results mirror those of the domain scores but more moderately. %, with our method demonstrating superior performance in generating well-grounded summaries.
Our method improves the groundedness score of both models by over 7\% compared to greedy search and by 5 to 3\%\ compared to beam search. As for the relevance score, we boost it by 7\% compared to greedy search in the case of the Phi-3 model. These improvements indicates that generations based on our decoding process contain slightly less hallucinated content.

\subsection{Summarization}\label{sec:eval_summarization}
To %evaluate the summarization capabilities of our method and 
compare our approach to state-of-the-art techniques, we evaluate its performance on generating the BHC section of the discharge summary of a MIMIC-III's patient admission, a well known task in the literature of clinical summarization \cite{searle_discharge_2023,adams_speer_2024}. We will define this task as the BHC task. We evaluate it on the $\Omega$ test set. As our method only requires a set of ontology classes to define a domain, we extended our approach to consider the BHC section as a domain. Just as presented in section \ref{sec:domain_analysis}, we performed a domain adaptation analysis on the BHC section of 1000 admissions to build our DCF dictionary. %This way, we aim to evaluate how versatile our method can be in the definition of domains (fields, text sections, ...). 
\paragraph{Baselines.} We compare our method against a standard single-pass generation using a predefined prompt 
%(detailed in \ref{appendix:bhc}) 
and other state-of-the-art techniques that underwent fine-tuning such as SPEER \cite{adams_meta-evaluation_2023} and \cite{searle_discharge_2023}'s dual Transformer.
Following the literature, we compute the ROUGE score as it is the most used for this task. 

\paragraph{Metrics.} Similarly to \cite{adams_speer_2024}, we also compute an hallucination score which measures how much the generated summary tends to mention concepts that are not mentioned in the original notes. Given the set $N$ of concepts in the clinical notes and the set $S$ of concepts in the generated summary, this score is defined as : 
\begin{equation}
    \text{HS} = \frac{|S - N|}{|S|} 
\end{equation}
However, this score is not entirely perfect since it does not take into account the uncertainty of the annotator used to tag the concepts. We thus compute the adjusted hallucination score (AHS) which incorporates the concepts from the reference summary into the metric. Given the set $R$ of concepts in the reference summary, this metrics is defined as :
\begin{equation}
    \text{AHS} = \frac{|S - (N \cup R)|}{|S|}
\end{equation}
The AHS will give room for concepts that can be inferred from the notes.
\begin{table*}[h]
    \centering
\begin{tabular}{clccccc}
\toprule
Fine-tuned                  & Method                     & \;\;R1$\uparrow$\;\;             & \;\;R2$\uparrow$\;\;            & \;\;RLSum$\uparrow$\;\;          & \;\;HS (\%) $\downarrow$\;\;        & \;\;AHS (\%) $\downarrow$\;\;       \\ \hline
\multirow{3}{*}{\cmark}    & Zephyr \cite{adams_speer_2024}                     & 25.00          & 6.90          & -              & -              & -              \\ \cline{2-7} 
                             & SPEER \cite{adams_speer_2024}             & \textbf{25.90}          & 7.10          & -              & -              & -              \\ \cline{2-7} 
                             & Dual Transformer \cite{searle_discharge_2023}                 & -              & \textbf{11.50}         & \textbf{34.90}          & -              & -              \\ \midrule
\multirow{3}{*}{\xmark} & Baseline (Phi-3)           & 26.59          & 4.75          & 13.85          & 45.70& 38.68\\ \cline{2-7} 
                             & Ours (Extraction)         & 25.94          & 4.70          & 13.30          & 38.47& 33.54\\ \cline{2-7} 
                             & Ours (Extraction+Pruning) & \textbf{28.41} & \textbf{5.47} & \textbf{13.93} & \textbf{37.95}& \textbf{33.08}\\ \midrule
\multirow{3}{*}{\xmark} & Baseline (Zephyr)          & 20.32          & 4.01          & 11.56          & 45.43& 41.60\\ \cline{2-7} 
                             & Ours (Extraction)         & 20.01          & 3.10          & 11.24          & 45.00& 41.18\\ \cline{2-7} 
                             & Ours (Extraction+Pruning) & \textbf{23.44} & \textbf{4.20} & \textbf{12.78} & \textbf{43.76}& \textbf{40.39}\\ \midrule
\end{tabular}
    \caption{Results for the BHC summarization task}
    \label{tab:results_bhc}
\end{table*}

The results shown in Table \ref{tab:results_bhc} show that our prompt-based method does not outperform fine-tuned models overall. Both the extraction and pruning step are needed to achieve the best performance. Likewise, our technique showed improvements in the hallucination and ajusted hallucination scores for the Phi-3 model. This effect is not as important with Zephyr. Possible explanations for these results include that the prompt structure was specifically tailored to maximize Phi-3's and that Zephyr has a higher number of parameters.

\section{Discussion and Limitations}
\subsection{Domain adaptation}
The results in Table \ref{tab:domain_score} suggest that our approach is really effective on its intended application. It is more efficient at producing more domain-adapted summaries than simply prompting the model to do so. These findings underscore the significance of the pruning step as it is the main step performing the domain adaptation. This also shows that the set of relevant concepts for a domain can be inferred from the data%without necessitating expert interventions
, as the pruning step is completely based on the initial domain adaptation analysis. These findings also implicitly confirm that our ontology-based decoding process improves model performance. Indeed, the effectiveness of the pruning step heavily relies on the values in the CSR which are determined using our custom decoding process. Furthermore, our method significantly enhances the summarization process interpretability by separating the extraction and adaptation phases, allowing for a more modular, interpretable process. This transparency allows for easier verification and validation of the generated content. Once the extraction phase is performed, the pruning phase can be used to adapt to any domain without the need to repeat the extraction. 
\subsection{Reduction of Hallucinations}
As shown in Table \ref{tab:results_bhc} by the hallucination score (HS) and adjusted hallucination score (AHS), our methodhandles  hallucinations better in the generated summaries compared to the baselines, especially with Phi-3. While the extraction phase seems to have a large impact on the performance of Phi-3, in the case of Zephyr, it is the pruning phase that creates a bigger jump in performance. The improvements of our method shown in Table \ref{tab:groundedness_score} also suggest that summaries generated using our custom ontology-guided decoding process are more likely to be factually consistent. This improvement in factuality and relevance is especially pronounced in the case of the Phi-3 model. %with an average improvement of 8\%. % These findings indicate that the model hallucinates less when asked about a certain concept, confirming the effectiveness of the scores computed in section \ref{sec:constrained_decoding}. 
Grounding each beam with the input and restraining the decoding process using an ontology contributes to hallucination reduction especially for a smaller model. %We can observe that the relevance scores of Zephyr's answers are way higher than Phi-3's. This is mainly due to Zephyr answering 'N/A' more frequently when prompted to summarize, thus being more sure when answering.

\subsection{Summarization Performance}
As Table \ref{tab:results_bhc} suggests, our technique demonstrates slight improvements in the summarization performance for clinical notes while providing a structured version of the same summary. This format is easier to query for clinicians. While this is not the best task to evaluate our method as it requires formulations specific to BHC summaries, it allows us to compare our results to state-of-the-art methods. Additionally, we also show that our definition of domains can be generalized since it is applicable to other tasks such as the BHC task. Finally, is is not surprising that our results did not outperform fine-tuned models since we used a prompting-based approach on models that were not specifically pre-trained to handle medical data.

%Besides, we outperforms some of these techniques like SPEER \cite{adams_speer_2024} in the ROUGE-1 metric despite not being fine-tuned on the MIMIC dataset. While most of the improvement is due to the Phi-3 model being better at summarization than Zephyr, we can notice that compared to the baseline, our method still improves the R1 score by almost 2 points in the case of the Phi-3 model. This suggest that our pruning phase is efficiently adapting the summary for the BHC section. The lower scores in R2 and RLSum can be explained by the fact that our models have not gone through a fine-tuning phase. Indeed, BHC summaries have a large variability in their format and content, which makes the summarization task pretty hard without any fine-tuning. We still see an improvement in the R2 score and RLSum compared to the baseline. %While the main focus of our method is to perform domain adaptation, we see that it can still be used to generate text sections without a degradation in summarization performance. Additionnally, 
% Plus, compared to other methods, we also outputs a structured summary. 
%An example of a structured summary is shown in section \ref{results_acess}.

\subsection{Limitations}
Since our approach relies on summarizing the notes around multiple concepts, the primary challenge is the computational overhead associated with the inference passes required by this method. While, they can easily be parallelized across all clinical notes and ontology classes, the overhead associated to beam search on all these passes poses challenges for scalability and deployment. Plus, our method is highly sensible to hyperparameters (prompt format, $k$, $\alpha$, ...) which makes it hard to optimize. Additionally, it heavily relies on a good annotator. % as it needs to be able to link words to ontology classes efficiently. 
 While such an annotator exists for SNOMED-CT, it might not be available for other ontologies. Finally, a significant limitation of our work is the absence of human assessment and gold standards with domain-related summaries. Due to time constraints in accessing medical experts, our evaluation only relies on automated evaluation methods which may not be always reliable. For instance, the NLI model employed to detect hallucinations in our generated content may not be optimal, as it lacks specific training on medical data. Consequently, the scores about hallucinated content may not be always trustworthy. Our future work will provide a qualitative evaluation of our results. %This potentially impacts the clinical validity and applicability of our results.

\section{Conclusion}

In this paper, we introduced a novel approach capable of generating domain-adapted summaries. We leveraged an ontology-guided decoding process to improve the factuality and relevance of the summarization process. We showed that guiding the model towards ontological concepts creates better domain-adapted summaries. Furthermore, our method improves performance in summarization, domain-adaptation, and groundedness on text input while providing a structured version of the summary. This structured version can easily be queried to extract relevant information in any use cases.

Applied to the medical domain, our approach highlights the potential for generating tailored summaries across various medical fields. While computational overhead remains a challenge, %the benefits of our method present a promising direction for addressing the documentation burden in healthcare settings. By providing a structured, adaptable, and more reliable approach to generating summaries from electronic health records, 
this work represents a step towards reducing clinician burnout and improving patient care through more efficient information synthesis.

\subsubsection*{Acknowledgements.} We gratefully acknowledge the financial support provided by the FRQS (Fonds de Recherche du Québec – Santé) Chair in Artificial Intelligence and Digital Health. This research also benefited from the computational resources and GPU infrastructure provided by the Digital Research Alliance of Canada (DRAC) and Calcul Québec. We extend our sincere thanks to Prof. John Kildea for his valuable input during initial discussions on the challenge of adapting summaries for specialists. 
\clearpage
% \bibliography{custom}
\bibliographystyle{splncs04}
\bibliography{custom}
\end{document}